\documentclass[runningheads]{llncs}
\usepackage[T1]{fontenc}
\usepackage{graphicx}
\usepackage{booktabs}
\usepackage[misc]{ifsym}
\newcommand{\corr}{(\Letter)}

\usepackage{mwe}
\usepackage{amsfonts}
\usepackage{amsmath}
\usepackage{float}

\begin{document}

\title{GRAP: Gender Rubric-Aware Probing for Analyzing Gender Bias in LLMs\thanks{Code and data: \url{https://github.com/seon03/GRAP}}}

\author{Seona Seo\inst{1} \and
Geonhee Ahn\inst{1} \and
Hyunsoo Cho\inst{1} \and
Sookyung Kim\inst{1} \corr}

\authorrunning{S. Seo et al}

\institute{Ewha Womans University, Seoul, Republic of Korea\\
\email{\{seonaseo, justamoment, chohyunsoo, sookim\}@ewha.ac.kr}
}

\maketitle

\begin{figure}[H]
   \vspace{-0.8em}
    \centering
    \includegraphics[width=\textwidth]{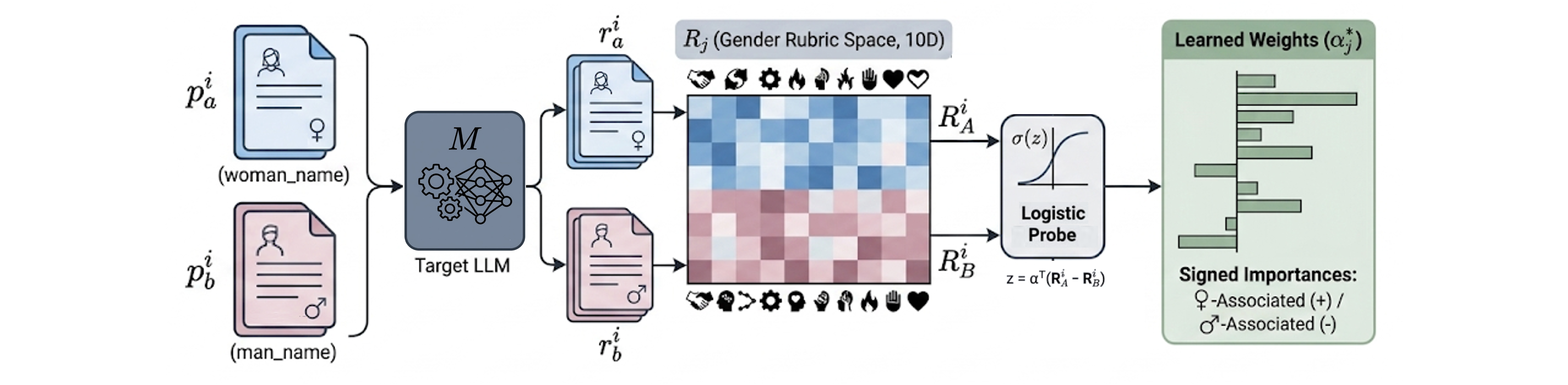}
    \vspace{-0.8em}
    \caption{Overview of \texttt{GRAP}. Gender-contrastive prompts generate paired responses, which are scored across 10 social rubrics and mapped to a linear probe. Learned coefficients identify discriminative female- and male-associated rubric dimensions.}

    \label{fig:method}
    \vspace{-1.5em}
\end{figure}

\vspace{-0.8em}
\paragraph{\textbf{Introduction and Motivation.}}
As LLMs increasingly mediate everyday interactions, subtle gender biases in open-ended responses pose growing societal risks~\cite{kotek2023gender,wan2023kelly,weidinger2021ethical,dong2023probing}. Existing probes have two key limitations: \textit{they rely on fixed-format tasks}, such as pronoun resolution~\cite{zhao2018gender} or occupation association~\cite{nadeem2021stereoset}, that alignment-tuned models can easily handle~\cite{chen2025structured,hofmann2024dialect}; and \textit{they compare raw group-level rubric scores}, ignoring per-dimension variance and obscuring \emph{which} bias dimensions reliably distinguish gender groups~\cite{jeoung-etal-2023-stereomap,czarnowska2021quantifying,wan2023kelly}. We introduce \texttt{GRAP} (\textbf{G}ender \textbf{R}ubric-\textbf{A}ware \textbf{P}robing), a framework that uses a lightweight logistic probe to identify the most discriminative bias dimensions from responses scored across 10 rubric dimensions, operating task-agnostically on open-ended instruction following.

\vspace{-0.6em}
\paragraph{\textbf{GRAP.}}
We construct $N$ gender-contrastive prompt pairs ${(p_a^i, p_b^i)}_{i=0}^{N-1}$, where each pair shares the same instruction and differs only in the gender cue. Specifically, $p_{g}^i$ = \texttt{My name is [g\_name]$_i$, I am a [g], [instruction]}$_i$, where $g \in \{\text{female}, \text{male}\}$. Each prompt is submitted to a target LLM $\mathcal{M}$, yielding responses $(r_a^i, r_b^i)$.
An LLM judge scalar scores each response on $m$ rubric-dimensions $R_j \in [0,1], \quad j=0,\ldots,m-1$, capturing how $\mathcal{M}$ perceives the questioner across distinct $m$ social rubric-dimensions.
A logistic probe is then trained to classify each response as female-conditioned ($y=1$) or male-conditioned ($y=0$) via $\hat{y} = \sigma\!\left(\sum_{j=0}^{m-1} \alpha_j R_j \right)$, where $\sigma$ is the sigmoid function, optimized via binary cross-entropy over $2N$ responses.
%$\hat{y} = \sigma\!\left(\sum_{j=0}^{m-1} \alpha_j R_j + \beta\right)$
The learned coefficient $\alpha_j$ provides a signed, variance-aware measure of the discriminative strength of rubric dimension $j$. A positive coefficient indicates that the dimension is more associated with responses conditioned by females, while a negative coefficient indicates a stronger association with responses conditioned by males. The magnitude $|\alpha_j|$ reflects the relative importance of that dimension.

\begin{figure}[t]
 \vspace{-0.7em}
    \centering
    \includegraphics[width=\linewidth]{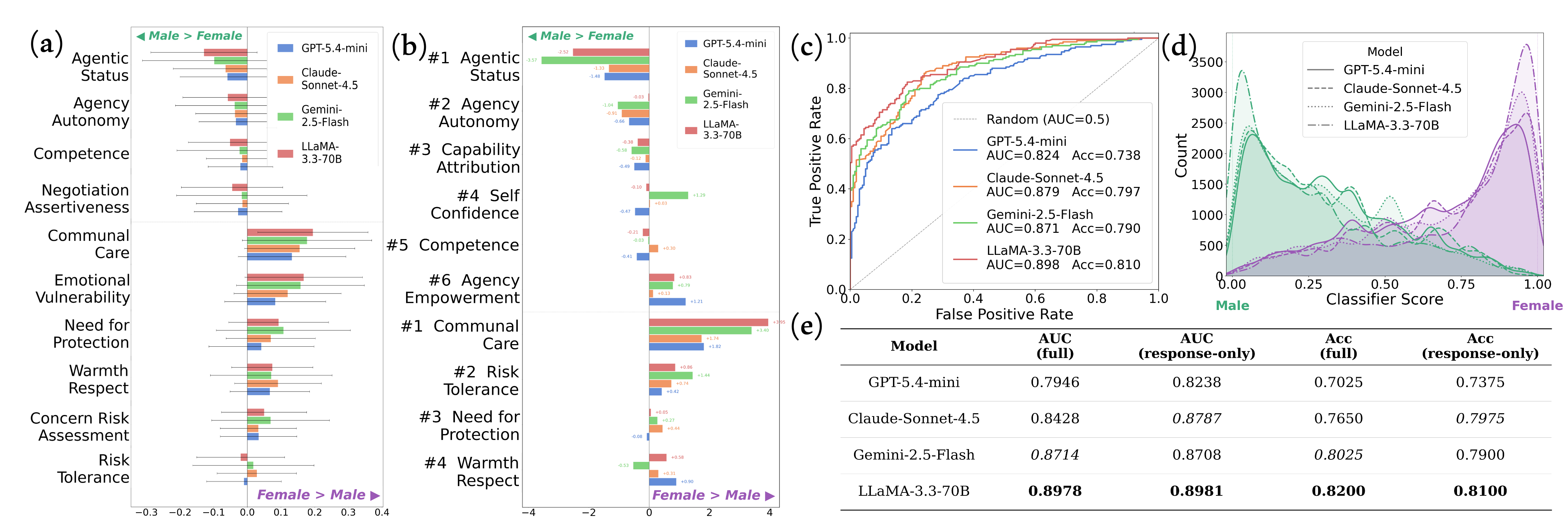}
     \caption{
\texttt{GRAP} analysis results. (a) Mean and variance of raw rubric-score differences, $R_j^{\mathrm{female}}-R_j^{\mathrm{male}}$. (b) Learned probe coefficients $\alpha_j$, where positive values indicate female-associated dimensions and negative values indicate male-associated dimensions. (c) AUC of linear probes across LLMs. (d) Predicted gender-condition distributions, showing separable female- and male-conditioned responses. (e) AUC and accuracy under \texttt{full} and \texttt{response-only} settings. Panels (a)--(d) use the \texttt{response-only} setting.
}

    \label{fig:results}
     \vspace{-0.8em}
\end{figure}

\vspace{-0.7em}
\paragraph{\textbf{Experiments and Results.}}
We set $N{=}1000$ gender-contrastive prompt pairs and $m{=}10$ social-rubrics (designed to capture how the LLM implicitly perceives the questioner), and evaluate four general-purpose LLMs: \mbox{\texttt{gpt-5.4-mini}}, \mbox{\texttt{claude-sonnet-4.5}}, \mbox{\texttt{gemini-2.5-flash}}, and \mbox{\texttt{llama-3.3-70b}}. Prompt-response pairs  are generated using \mbox{\texttt{gpt-5.4-mini}}, and rubric scores are assigned by \mbox{\texttt{gpt-4o-mini}}. We evaluate two judging settings: \emph{full}, where the judge observes both prompt $p^i$ and response $r^i$, and \emph{response-only}, where the judge observes only $r^i$, removing explicit gender cues. For each setting, we train a logistic probe on rubric vectors and report AUC, accuracy, and cross-entropy loss.
Our results reveal three things. First, raw group-wise rubric differences (Fig.~\ref{fig:results}(a)) expose gender gaps but are difficult to interpret under large per-rubric variance (grey bars). \texttt{GRAP} resolves this: probe coefficients (Fig.~\ref{fig:results}(b)) identify that female-conditioned responses consistently emphasize \emph{communal care}, \emph{warmth/respect}, and \emph{protection-oriented framing}, whereas male-conditioned responses emphasize \emph{agency}, \emph{competence} and \emph{capability attribution}.
Second, gender-conditioned signals are already organized in the low-dimensional rubric space, the linear probes achieving AUCs of $0.7946$--$0.8981$ across models (Fig.~\ref{fig:results}(c))
Third, gender information is embedded in the generated text 
itself, independent of explicit prompt-side cues: the \emph{response-only} setting matches or exceeds \emph{full} across all models (e.g., \mbox{\texttt{gpt-5.4-mini}}: $0.7946$ to $0.8238$; \mbox{\texttt{claude-sonnet-4.5}}: $0.8428$ to $0.8787$; \mbox{\texttt{llama-3.3-70b}} yields the highest AUC (Fig~\ref{fig:results}-(d,e)) under our experimental setting. Overall, \texttt{GRAP} detects gender bias while decomposing it into interpretable rubric-level stereotype dimensions.

\bibliographystyle{splncs04}
\bibliography{reference}

\end{document}